# Performance Evaluation of t-SNE and MDS Dimensionality Reduction Techniques with KNN, ENN and SVM Classifiers

Shadman Sakib
Department of Electrical and Electronic Engineering
IUBAT, Dhaka-1230, Bangladesh
sakibshadman15@gmail.com

Md. Abu Bakr Siddique
Department of Electrical and Electronic Engineering
IUBAT, Dhaka-1230, Bangladesh
absiddique@iubat.edu

Md. Abdur Rahman
Department of Electrical and Electronic Engineering
IUT, Gazipur-1704, Bangladesh
marahman.iut@gmail.com

*Abstract*—The central goal of this paper is to establish two commonly available dimensionality reduction (DR) methods i.e. t-distributed Stochastic Neighbor Embedding (t-SNE) and Multidimensional Scaling (MDS) in Matlab and to observe their application in several datasets. These DR techniques are applied to nine different datasets namely CNAE9, Segmentation, Seeds, Pima Indians diabetes, Parkinsons, Movement Libras, Mammographic Masses, Knowledge, and Ionosphere acquired from UCI machine learning repository. By applying t-SNE and MDS algorithms, each dataset is transformed to the half of its original dimension by eliminating unnecessary features from the datasets. Subsequently, these datasets with reduced dimensions are fed into three supervised classification algorithms for classification. These classification algorithms are K Nearest Neighbors (KNN), Extended Nearest Neighbors (ENN), and Support Vector Machine (SVM). Again, all these algorithms are implemented in Matlab. The training and test data ratios are maintained as ninety percent: ten percent for each dataset. Upon accuracy observation, the efficiency for every dimensionality technique with availed classification algorithms is analyzed and the performance of each classifier is evaluated.

## I. INTRODUCTION

Dealing with huge data comprises key difficulties of information technology in the new century. Real world data appears high-dimensional [1] such as images, sound signals comprising different dimensions to denote data that is challenging to analyze. Data of higher dimensions are increasingly overwhelming to define and control the networks between concepts. Therefore, to manage such real-life data sufficiently, its dimensionality ought to be reduced. Dimensionality reduction (DR) is a system exploited for diminishing complications for evaluating high dimensional data. Numerous methods are available, such as some conventional techniques principal component analysis (PCA) [2], multidimensional scaling (MDS), self-organizing map (SOM), etc., delivering a visual information assessment analyzer that is being used extensively throughout different domains, for example, social sciences or bioinformatics for years. They reduce the dimensions from the data of the original information. DR can often be additionally fragmented into feature selection and feature extraction. One does it to eliminate an input function of an original statement, another is to decrease the dimensionality of a previously removed high-dimensional function [3],[4]. It is important to transform on to a relatively small and progressively appropriate data collection when splitting up a large data set of higher dimensions [5].

t-SNE is a remarkable nonlinear DR technique that is employed to visualize data with a higher dimension [6] by giving a position to each set of data inside a 2D or 3-dimensional map. Shaham et al. [7] demonstrated a theoretical analysis of t-SNE where they mathematically demonstrated that the structure of the loss function of SNE indicates that the entire family maps all around separated clusters in a quantitative manner. Density-based representation of the t-SNE implantation has been used in numerous projects although it is slow in measuring the offline techniques [8]. Despite t-SNE has made itself into several life science applications, in recent times, it has demonstrated tremendous impact in the medical sector. In addition, t-SNE is known for analyzing single-cell genome sequence data and is extensively applied nowadays to genetic data [9].

On the other hand, MDS is one of the prominent non-linear dimensionality reduction methods used to analyze the proximity (similarity or dissimilarity) of data by reducing the data into a low dimensional space. Standard MDS processes become sufficient in situations where the data is particularly nonmetric or scattered. Analysis of nonlinear DR procedures for breast MRI segmentation has been discussed in [10] where they utilized nonlinear DR methods for evaluating the performance and MDS was one of the approaches among them. Touati et al. [11] exhibited a proficient MDS reliant on a multi-axial dimension for the acknowledgment of human activity in a video sequence. Moreover, few mechanisms were evolved over the past 15 years to allow substantial-scale applications to implement the classical MDS algorithm. The continuous analysis integrates MDS for rendering high-dimensional sets of data and a significant decrease in linear dimensionality.

For this study, we utilized 9 UCI machine learning repository collection of data [12] to implement KNN, ENN, and SVM algorithms for the DR technique t-SNE and MDS and made the comparison between the performances. For each dataset, variations of accuracies such as F-measure and G-mean values of the algorithms were observed to allow comparison in the performances of the algorithms. The simulation results were acquired by using Matlab. The article overview is as follows. The algorithms overview and mechanism of the experimental system were shortly explained in section II. Section III discusses the result of the experiment and subsequently, section IV concludes the paper.

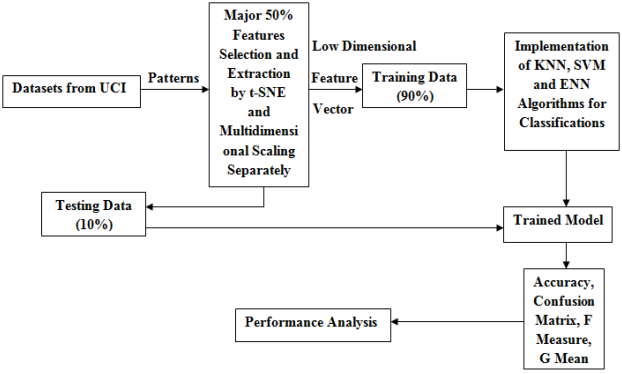

Fig. 1. Workflow illustrating the entire procedure used on this paper for the performance analysis of t-SNE and MDS techniques with supervised classifiers

## II. ALGORITHMS' OVERVIEW AND MECHANISM

### A. Dimensionality Reduction Techniques Algorithm

*1) t-distributed stochastic neighbor embedding (t-SNE)*

t-SNE [6] is one of a very few algorithms that are suitable for the simultaneous preservation of data recollecting both neighborhood and global framework of the data. Assuming a set of high-dimensional objects $N = \{x_1, x_2, ..., x_N\}$ and a function $d(x_i, x_j)$. The Euclidean distance is $d(x_i, x_j) = \|x_i - x_j\|$. t-SNE will first calculate the conditional probabilities $p_{j|i}$ of the between the similar objects of data points $x_i$ and $x_j$. Mathematically conditional probability $p_{j|i}$ given by,

$$p_{j|i} = \frac{\exp(-\|x_i - x_j\|^2 / 2*\sigma_i^2)}{\sum_{k \neq i} \exp(-\|x_i - x_k\|^2 / 2*\sigma_i^2)} \quad .........(1)$$

Now an identical conditional probability $q_{j|i}$ is computed for the low-dimensional set of data $y_i$ and $y_j$ assigning to the high-dimensional set of data $x_i$ and $x_j$, given by,

$$q_{j|i} = \frac{\exp(-\|y_i - y_j\|^2)}{\sum_{k \neq i} \exp(\|y_i - y_k\|^2)^2} \quad .........(2)$$

Again for exhibiting similar pair of data, we set $q_{i|i} = 0$. As Kullback-Leibler divergences minimized by t-SNE between P and Q, the amount of the cost function C is given by,

$$C = KL(P \| Q) = \sum_{i \neq j} p_{ij} * \log \frac{p_{ij}}{q_{ij}} \quad .........(3)$$

Therefore, the cost function is implemented using a type of gradient descent in (3). The gradient can be shown in a simplified form as,

$$\frac{\partial C}{\partial y_i} = 4 \sum_{j \neq i} (p_{ij} - q_{ij}) * q_{ij} * Z(y_i - y_j) \quad .........(4)$$

*2) Multidimensional Scaling (MDS)*

Multidimensional Scaling (MDS) is a multidimensional non-linear DR method that converts present high-dimensional data into shallower-dimensional data by retaining pair distance information in low-dimensional data space [13]. It takes a dissimilarity matrix $D = [d_{ij}]$ with original dimension m, between pair of items, in which $d_{ij}$, the length among i and j. The expected output matrix is X with reduced dimension d (usually d = 1, 2, or 3) whose design minimizes a loss function known as strain [14]. The strain is defined as,

$$Strain_D(x_1, x_1, ......, x_N) = \left( \frac{\sum_{i,j}(b_{ij} - \langle x_i, x_j \rangle)^2}{\sum_{i,j} b_{ij}^2} \right)^{1/2} \quad .........(5)$$

MDS is based on the fact that the coordinate matrix X can be obtained by eigen-decomposition from $B = X*X'$. Again, the matrix B can be calculated from the dissimilarity matrix D by applying double centering [15]. The following steps will characterize the Classical MDS algorithm.

1. Evaluating square dissimilarity matrix $D^2 = [d_{ij}^2]$
2. Compute centering matrix $J = I - \frac{1}{n}11'$, in which n is the actual number of items.
3. Compute $B = -\frac{1}{n} JD^2 J$
4. Determine the d largest eigenvalues $(\lambda_1, \lambda_2, \lambda_3, ......, \lambda_d)$ and corresponding eigenvectors $(e_1, e_2, e_3, ......, e_d)$ of matrix B.
5. Finally, $X = E_d \Lambda_d^{1/2}$, where $E_d$ is the matrix of d eigenvectors and the diagonal matrix of d ($\Lambda_d$) eigen-values of matrix B

### B. Classification Algorithms'

*1) K-Nearest Neighbors (KNN)*

KNN [16] is a specific instance-based indolent computing classifier used to predict a new categorization of test points in a database where the data items are split onto a range of classes. Figure 2 demonstrates the basic KNN classifier diagram.

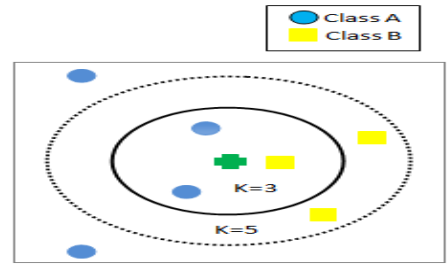

Fig. 2. Basic KNN classifier structure. The test sample (i.e., green rectangular prism) is assigned to the blue circle class if k = 3 while for k = 5 it gets allocated to the yellow squared class

A KNN classifier measures conditional probability for each group in a particular set of nearest neighbors, k, an arbitrary test point, x, and a distance parameter, d. And at last, the unspecified test point, x gets classified with optimum probability for the class.

$$P(z=w \mid X=k) = \frac{1}{K} \sum_{i \in M} I(z^{(i)} = w) \quad \ldots\ldots (6)$$

*2) Extended Nearest Neighbors (ENN)*

ENN uses detailed classification wise details of all input data to be trained through complete data distribution, thus improving efficiency in categorization [17]. Let all the detailed classification wise data point $Q_i$ for class i containing the aggregate measurements $Z_1$ and $Z_2$ can be evaluated for every classification with its nearest neighbors as,

$$Q_i = \frac{1}{n_i k} \sum_{x \in Z_i} \sum_{r=1}^{k} I_r(x, Z = Z_1 \cup Z_2); i = 1,2\ldots\ldots(7)$$

Depending on the target component, the ENN then classifies the sample as follows,

$$\beta_{ENN} = \arg\max_{j \in 1,2} \sum_{i=1}^{2} Q_i{}^j = \arg\max_{j \in 1,2} \Theta^j \ldots\ldots(8)$$

Then the two-class ENN framework can conveniently be adapted towards multi-class categorization by,

$$\beta_{ENN} = \arg\max_{j \in 1,2\ldots\ldots N} \sum_{i=1}^{N} Q_i{}^j \ldots\ldots(9)$$

*3) Support Vector Machine (SVM)*

The SVM algorithm incorporates linear categorization through accomplishing the hyperplane which often widens the distance in both two categories. Support vectors are the data points or independent variables that determine the hyperplane [18]. The essential design of the SVM classifier illustrated in Figure 3.

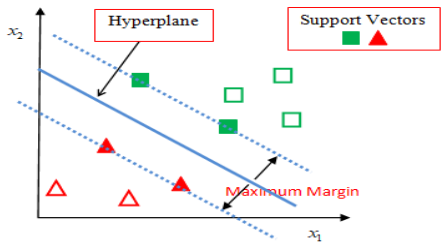

Fig. 3. Strategy representing the SVM classifier along with its hyperparameters

Over a particular training data of n objects in the form $(\vec{\alpha}_1, \beta_1),\ldots,(\vec{\alpha}_n, \beta_n)$ wherein $\beta_i = \pm 1$ the class corresponding to the series of points $\vec{\alpha}_i$ is expressed. The number of points $\vec{\alpha}_i$ can be seen on a hyperplane as,

$$\vec{u} \cdot \vec{\alpha}_i + b = 0 \ldots\ldots(10)$$

Although hyperplane is manifested by the drawback of two classes $\beta_i = \pm 1$, margin P is twice the length to those of the neighboring circumstances,

$$P = \frac{2}{\|\vec{u}\|} \ldots\ldots(11)$$

Lastly, the concern to refine P is adequate to maximize the problem $\|\vec{u}\|$. To address this problem, the SVM classification system [19] provides the solution to the subsequent problem of optimization,

$$\arg\max_{u,b,\xi} \frac{1}{2} u \cdot u + C \sum_{i=1}^{n} \xi_i \ldots\ldots(12)$$

## III. RESULT ANALYSIS AND DISCUSSION

We utilized nine datasets from the UCI machine learning repository throughout this experiment to assess the performance of both the DR technique (t-SNE and MDS) using the supervised classification algorithms KNN, ENN and SVM and then made a comparison among the DR techniques. The classification accuracies for the execution DR technique t-SNE is illustrated in Figure 4.

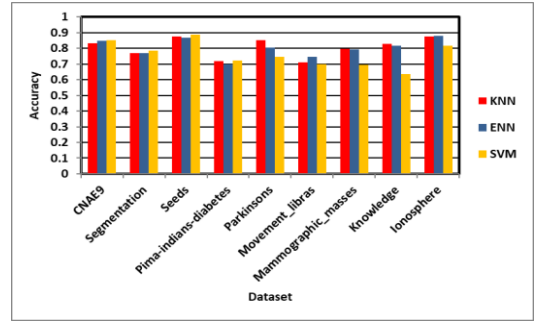

Fig. 4. Classification accuracy plot for the DR technique t-SNE

From the observation, it is noticeable that, in the case of t-SNE, both of the SVM and KNN works well with the data sets as regards classification accuracy. The SVM classifier produced the best accuracy (0.8876) for Seeds. After using t-SNE, KNN yields significant changes over SVM on a few datasets (e.g., Parkinsons, Mammographic masses, Knowledge). ENN showed compatible results for almost all the datasets having the highest accuracy (0.8774) for Ionosphere dataset. Besides, ENN performed better than SVM as well as KNN for the Movement Libras dataset. The performance evaluation metrics for all of the data sets used with the classifiers (KNN, ENN, and SVM) incorporated by t-SNE as the DR method is demonstrated in Table I.

TABLE I. PERFORMANCE EVALUATION WITH T-SNE

| Data set | F-measure | | | G-mean | | |
|---|---|---|---|---|---|---|
| | KNN | ENN | SVM | KNN | ENN | SVM |
| CNAE9 | 0.8295 | 0.8497 | 0.8502 | 0.8334 | 0.8531 | 0.8534 |
| Segmentation | 0.7602 | 0.7611 | 0.7775 | 0.7707 | 0.7686 | 0.7852 |
| Seeds | 0.8667 | 0.8602 | 0.8814 | 0.8727 | 0.8651 | 0.8856 |
| Pima-indians-diabetes | 0.6784 | 0.6776 | 0.6855 | 0.6808 | 0.6810 | 0.6876 |
| Parkinsons | 0.7781 | 0.7455 | 0.4255 | 0.7878 | 0.7548 | 0.4310 |
| Movement_libras | 0.7048 | 0.7378 | 0.9630 | 0.7181 | 0.7469 | 0.7041 |
| Mammographic_masses | 0.7937 | 0.7915 | 0.6881 | 0.7945 | 0.7924 | 0.6900 |
| Knowledge | 0.7966 | 0.7938 | 0.5469 | 0.8038 | 0.7992 | 0.5566 |
| Ionosphere | 0.8501 | 0.8596 | 0.7868 | 0.8561 | 0.8614 | 0.7899 |

In contrast, in the case of MDS, Table II demonstrates the resulting details. We reduced dimensionality by half using t-SNE and MDS. Both the performance evaluation metrics (F measure and G mean) values signify that SVM demonstrates good outcomes if the MDS technique is implemented. Conversely, on a few data sets, the KNN has an advantage across ENN and SVM in using t-SNE.

TABLE II. PERFORMANCE EVALUATION WITH MDS

| Data set | F-measure | | | G-mean | | |
|---|---|---|---|---|---|---|
| | KNN | ENN | SVM | KNN | ENN | SVM |
| CNAE9 | 0.8696 | 0.9047 | 0.9361 | 0.8732 | 0.9070 | 0.9374 |
| Segmentation | 0.7515 | 0.7529 | 0.8767 | 0.7619 | 0.7604 | 0.8816 |
| Seeds | 0.8723 | 0.8673 | 0.9024 | 0.8778 | 0.8731 | 0.9059 |
| Pima-indians-diabetes | 0.6579 | 0.6532 | 0.6830 | 0.6604 | 0.6565 | 0.6904 |
| Parkinsons | 0.7640 | 0.7321 | 0.7813 | 0.7743 | 0.7415 | 0.7967 |
| Movement_libras | 0.6810 | 0.7398 | 0.6581 | 0.7012 | 0.7498 | 0.6754 |
| Mammographic_masses | 0.8031 | 0.8010 | 0.8145 | 0.8038 | 0.8019 | 0.8164 |
| Knowledge | 0.6874 | 0.6912 | 0.6228 | 0.6952 | 0.6979 | 0.6402 |
| Ionosphere | 0.8266 | 0.8656 | 0.8568 | 0.8376 | 0.8681 | 0.8610 |

The accuracy measurement among all the classifiers for the DR technique, MDS is depicted in Figure 5. The average effectiveness of the SVM is significantly greater than the nearest neighbor predicated algorithms, as demonstrated in Table II. With CNAE9 data set the highest classification accuracy (0.9383) was obtained. ENN thus outperforms the SVM and KNN algorithms on certain datasets (e.g., Movement Libras, Information, and Ionosphere). The ENN has the highest accuracy (0.8826) for the Ionosphere data set.

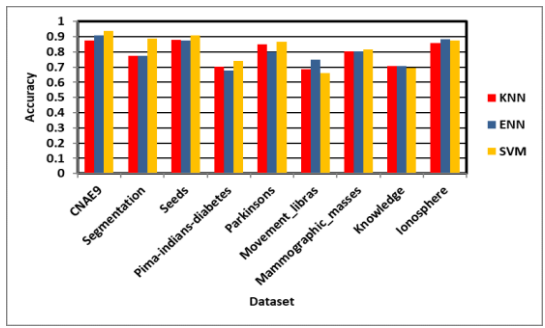

Fig. 5. Classification accuracy plot for the DR technique MDS

It must be noted that the performance of the KNN classifier is usually based on the k value, the number of nearest neighbors chosen in regards to the data set. The KNN classifier had demonstrated impressive performance for k=5 in this analysis. The results suggest that t-SNE is more compatible with nearest neighbor-based algorithms; however, MDS seems to be more consistent with the SVM technique.

## IV. CONCLUSION

For this analysis, we experimented with t-SNE and MDS dimensionality reduction techniques with supervised classification algorithms by KNN, SVM and ENN. All algorithms were validated from the UCI machine learning database on nine distinct data sets. After evaluating effectively, it is clear that SVM, and KNN with t-SNE work reasonably well on almost all datasets. SVM with t-SNE has given the highest accuracy of 88.76% on the Seeds dataset. However, ENN with t-SNE outperforms both SVM and KNN for Movement Libras dataset. Now, SVM with the MDS DR method generally outperforms nearest neighbor based classifiers and has demonstrated the highest classification accuracy of 93.83% on the CNAE9 dataset. Upon observation of F Measure and G Mean values, it is clear that in the case of t-SNE, KNN generally showed better performance over SVM and ENN classifiers; on the other hand, SVM demonstrated better performance when MDS is applied. These observations demonstrate that SVM is more suitable with the MDS method, while KNN and ENN are better compatible with t-SNE methods.

REFERENCES

[1] M. Song, H. Yang, S. H. Siadat, and M. Pechenizkiy, "A comparative study of dimensionality reduction techniques to enhance trace clustering performances," *Expert Syst. Appl.*, vol. 40, no. 9, pp. 3722–3737, 2013.

[2] S. Wold, K. Esbensen, and P. Geladi, "Principal component analysis," *Chemom. Intell. Lab. Syst.*, vol. 2, no. 1–3, pp. 37–52, 1987.

[3] M. Holmes, A. Gray, and C. Isbell, "Fast SVD for large-scale matrices," in *Workshop on Efficient Machine Learning at NIPS*, 2007, vol. 58, pp. 249–252.

[4] M. Siddique, A. Bakr, S. Sakib, M. Rahman, and others, "Performance Analysis of Deep Autoencoder and NCA Dimensionality Reduction Techniques with KNN, ENN and SVM Classifiers," *arXiv Prepr. arXiv1912.05912*, 2019.

[5] S. K. Joshi and S. Machchhar, "An evolution and evaluation of dimensionality reduction techniques—A comparative study," in *2014 IEEE International Conference on Computational Intelligence and Computing Research*, 2014, pp. 1–5.

[6] L. van der Maaten and G. Hinton, "Visualizing data using t-SNE," *J. Mach. Learn. Res.*, vol. 9, no. Nov, pp. 2579–2605, 2008.

[7] U. Shaham and S. Steinerberger, "Stochastic neighbor embedding separates well-separated clusters," *arXiv Prepr. arXiv1702.02670*, 2017.

[8] K. Shekhar, P. Brodin, M. M. Davis, and A. K. Chakraborty, "Automatic classification of cellular expression by nonlinear stochastic embedding (ACCENSE)," *Proc. Natl. Acad. Sci.*, vol. 111, no. 1, pp. 202–207, 2014.

[9] W. Li, J. E. Cerise, Y. Yang, and H. Han, "Application of t-SNE to human genetic data," *J. Bioinform. Comput. Biol.*, vol. 15, no. 04, p. 1750017, 2017.

[10] A. Akhbardeh and M. A. Jacobs, "Comparative analysis of nonlinear dimensionality reduction techniques for breast MRI segmentation," *Med. Phys.*, vol. 39, no. 4, pp. 2275–2289, 2012.

[11] R. Touati and M. Mignotte, "MDS-based multi-axial dimensionality reduction model for human action recognition," in *2014 Canadian Conference on Computer and Robot Vision*, 2014, pp. 262–267.

[12] A. Frank, A. Asuncion, and others, "UCI machine learning repository, 2010," *URL http//archive. ics. uci. edu/ml*, vol. 15, p. 22, 2011.

[13] A. Mead, "Review of the development of multidimensional scaling methods," *J. R. Stat. Soc. Ser. D (The Stat.*, vol. 41, no. 1, pp. 27–39, 1992.

[14] I. Borg and P. J. F. Groenen, *Modern multidimensional scaling: Theory and applications*. Springer Science & Business Media, 2005.

[15] F. Wickelmaier, "An introduction to MDS," *Sound Qual. Res. Unit, Aalborg Univ. Denmark*, vol. 46, no. 5, pp. 1–26, 2003.

[16] T. Cover and P. Hart, "Nearest neighbor pattern classification," *IEEE Trans. Inf. theory*, vol. 13, no. 1, pp. 21–27, 1967.

[17] M. M. R. Khan, R. B. Arif, M. A. B. Siddique, and M. R. Oishe, "Study and observation of the variation of accuracies of KNN, SVM, LMNN, ENN algorithms on eleven different datasets from UCI machine learning repository," in *2018 4th International Conference on Electrical Engineering and Information & Communication Technology (iCEEiCT)*, 2018, pp. 124–129.

[18] M. A. Hearst, S. T. Dumais, E. Osuna, J. Platt, and B. Scholkopf, "Support vector machines," *IEEE Intell. Syst. their Appl.*, vol. 13, no. 4, pp. 18–28, 1998.

[19] P.-Y. Hao, "Interval regression analysis using support vector networks," *Fuzzy sets Syst.*, vol. 160, no. 17, pp. 2466–2485, 2009.